\documentclass[sigplan,nonacm]{acmart}
\renewcommand\footnotetextcopyrightpermission[1]{}
\usepackage{array}
\usepackage{ragged2e}
\usepackage{tikz}
\usetikzlibrary{automata, positioning, calc, arrows.meta, fit}
\usepackage{mathpartir}
\usepackage[normalem]{ulem}
\usepackage{listings}
\usepackage{xspace}
\usepackage{multirow}
\usepackage{wasysym}
\usepackage{subcaption}
\usepackage{enumitem}
\usepackage{tcolorbox}
\usepackage{algorithm}
\usepackage[noend]{algpseudocode}
\tcbuselibrary{theorems}
\setlist[enumerate]{nosep}
\setlist[itemize]{nosep}
\usepackage{booktabs}
\usepackage{graphicx}
\usepackage{arydshln} %
\usepackage{siunitx}
\newcolumntype{L}[1]{>{\raggedright\arraybackslash\hspace{0pt}}p{#1}}
\newcolumntype{C}[1]{>{\centering\arraybackslash\hspace{0pt}}p{#1}}
\newcolumntype{R}[1]{>{\raggedleft\arraybackslash\hspace{0pt}}p{#1}}

\theoremstyle{definition}

\providecommand{\shuffle}{\parallel}

\newtcbtheorem[number within=section]{mytheo}{\small API Call}%
{colback=green!5,colframe=green!35!black,left=1mm,top=1mm,bottom=1mm}{th}

\definecolor{darkred}{rgb}{0.65,0.1,0.1}

\newcommand{\LLMguardrails}[0]{shieldagent, nemo, fedorov2024llamaguard3, kang2025csafegen, li2025piguard, shi2025promptarmor, llamafirewall, llamaguard}
\newcommand{\sys}{Sapien\xspace}

\newcommand{\eg}{{e.g.},\xspace}
\newcommand{\ie}{{i.e.},\xspace}
\newcommand{\one}{({\em i}\/)}
\newcommand{\two}{({\em ii}\/)}
\newcommand{\three}{({\em iii}\/)}

\newcommand{\llmjudge}[0]{shieldagent, nemo, fedorov2024llamaguard3, kang2025csafegen, li2025piguard, shi2025promptarmor, llamafirewall, llamaguard}

\definecolor{codegreen}{rgb}{0,0.4,0}
\definecolor{codegray}{rgb}{0.5,0.5,0.5}
\definecolor{codepurple}{rgb}{0.58,0,0.82}
\definecolor{backcolour}{rgb}{0.95,0.95,0.92}

\usepackage{sourcecodepro}
\usepackage[T1]{fontenc}
\lstdefinestyle{rust}{
    language=Python,
    commentstyle=\color{codegreen},
    keywordstyle=\color{codepurple},
    stringstyle=\color{blue},
    basicstyle=\ttfamily\scriptsize,
    breakatwhitespace=false,
    breaklines=true,
    captionpos=b,
    keepspaces=true,
    showspaces=false,
    showstringspaces=false,
    showtabs=false,
    tabsize=2,
    frame=single,
    rulecolor=\color{black!40},
    xleftmargin=3.4pt,
    xrightmargin=3.4pt,
    morekeywords={defer, deferred, PREDICATE, verify_with_ai, @},
}
\lstdefinestyle{inline-rust}{
    style=rust,
    frame=none,
    basicstyle=\ttfamily\footnotesize
}

\newcommand{\code}[1]{\lstinline[style=inline-rust]|#1|}

\begin{document}

\title{\sys: A Stateful Policy Engine\texorpdfstring{\\}{ }for Autonomous AI Agents}

\author{Corinn Tiffany$^{*,1}$, Wen Zhang$^{*,1}$, Eugene Bagdasarian$^{1,2}$, Lillian Tsai$^1$\vspace{0.2em}}
\affiliation{
\institution{$^1$Google, $^2$University of Massachusetts Amherst}
\city{}\country{}}

\begin{abstract}

Contextual security defenses prevent AI agents from taking rogue actions by synthesizing a task-specific policy and enforcing it on the agent's tool calls.
In multi-step tasks, however, which actions are valid often depends on what the agent has already done and learned.
We present \sys, a policy engine for enforcing \emph{stateful} contextual policies.
A \sys policy specifies permitted tool-call sequences using a regular expression extended with stateful predicates, deferred policy generation, and scoped semantic checks.
We show that \sys stays within a few percent of an unconstrained agent's utility.
Even if the agent is fully hijacked, \sys's policies rule out 93--95\% of attacks on AgentDojo and 62--85\% on Toolathlon (twice as many as tool allowlists on long-horizon tasks).
\end{abstract}

\maketitle
\pagestyle{plain}
\def\thefootnote{*}\footnotetext{Equal contribution. Correspondence to \href{mailto:wnz@google.com}{wnz@google.com}.}\def\thefootnote{\arabic{footnote}}

\section{Introduction}

AI agents use large language models~(LLMs) to autonomously plan and carry out natural-language tasks by calling external tools~\cite{agentdefn_google,agentdefn_nvidia,agentdefn_github}.
Unlike traditional programs governed by hard-coded control flow, agents rely on the LLM's reasoning to interpret tool outputs and dynamically plan their next steps, allowing them to tackle complex, long-horizon tasks.
Increasingly, developers aim for \emph{general-purpose agents}, equipping AI models with a broad suite of tools to handle open-ended tasks across domains on behalf of a user---for example, managing online accounts, organizing meetings and notes, or coordinating travel.

Yet delegating tool invocation to an LLM makes agents inherently unreliable and
prone to \emph{rogue actions}---tool calls that diverge from the user's intent.
As trajectories span more steps, agents are more likely to encounter untrusted
content that hijacks them via prompt injection~\cite{indirectpi, willisonpi}, or
to spontaneously drift from their goal, misinterpret a tool output, or improvise
an unintended workaround~\cite{meltdown}. In practice, agents have already
emptied user inboxes~\cite{techcrunch-email}, deleted production
databases~\cite{fortune-delete}, attacked external
infrastructure~\cite{huggingface_openai}, and executed attacker-controlled
commands~\cite{clinejection}, making this lack of reliability a primary obstacle
to deploying agents in production~\cite[\S7]{pan2026measuringagentsproduction}.

In classical systems, a common defense against rogue actions interposes a
reference monitor that checks each action against a
policy~\cite{saltzerschroeder, enforceablepolicies}. Such policies are typically
\emph{static}: if a program's purpose is known ahead of time, its appropriate 
actions can be specified in advance. For a general-purpose agent, however,
whether an action is legitimate depends on the user's task~\cite{conseca}:
invoking the \code{send_money} tool is appropriate when the user asks to
\emph{``refund the person who sent me money last week''}, but is a rogue action
when the user asks to \emph{``summarize my inbox.''} A static policy must either
permit \code{send_money} and leave the inbox summary vulnerable to theft, or
forbid it and break legitimate refunds.

Because agents' appropriate actions vary by task, a meaningful policy must be tailored to the task at hand.
We therefore adopt the framework of \emph{contextual agent security}~\cite{conseca}: for each request, a policy generator synthesizes a just-in-time, task-specific policy from the user's prompt, and a reference monitor enforces it over the agent's tool calls (\S\ref{s:setup:contextual}).
Although synthesizing a policy from natural language also requires an LLM, the
policy generator operates solely on the user's prompt and tool
documentation---isolated from untrusted runtime data and multi-step
execution---to most reliably produce a policy that reflects user intent.

Ideally, a task's policy should admit appropriate trajectories that accomplish
the task and reject all others. How closely a policy can approach this ideal is
bounded by the expressiveness of the policy language, and we argue that tightly
bounding an agent requires a \emph{stateful} policy language (\S\ref{s:setup:good}),
in which what the agent may do next depends on the actions it has already taken
and the data it has observed.

This paper presents \textbf{\sys}, a policy engine for AI agents based on a stateful policy language.
Following the contextual agent security framework described above, \sys synthesizes a policy in this language from the user's task prompt and tool documentation ahead of execution, and interposes on the agent's tool calls to enforce it (\S\ref{s:overview}).
A \sys policy specifies permitted sequences of tool calls using regular expressions---sequencing, choice, and repetition---and extends them with three mechanisms to incorporate runtime context (\S\ref{s:policy_lang}):
\begin{enumerate} %
    \item\label{item:pred} \textbf{Predicates over prior tool-call results}: Policies can bind return values of completed tool calls to variables, and constrain arguments against these variables via predicates (\eg requiring \code{send_money} to use a sender and amount from the read transaction history, \S\ref{s:lang:bind}).
    \item\label{item:defer} \textbf{Deferred policy generation}: When a prompt delegates a sub-task to an external source (\eg \emph{``complete the items on my TODO list''}), \sys places a \emph{hole} in the policy and expands that hole only after the agent fetches the designated source, avoiding a blanket permission for that sub-task (\S\ref{s:lang:defer}).
    \item\label{item:judge} \textbf{Semantic predicates}: For natural-language properties that cannot be checked by code (\eg whether an email faithfully summarizes a document), a predicate may pose a scoped yes/no question to an LLM judge over the specific values being checked, further tightening what the policy permits (\S\ref{s:lang:judge}).
\end{enumerate}

Enforcing such a policy over an agent's trajectory requires tracking the agent's
progress while leaving it free to choose among valid execution paths. \sys
compiles a policy into an augmented nondeterministic finite automaton
(\S\ref{s:system}): each state marks a point of progress through the policy, and
transitions check argument predicates, bind tool results to variables, and
expand deferred holes.
As the agent executes, \sys tracks the set of active
\emph{configurations}---automaton states paired with variable bindings---and
denies any tool call that no active configuration permits. 
\sys's enforcement only uses potentially untrusted data retrieved by the agent
during execution in ways prescribed by the policy, which is derived to reflect
user intent by design (\S\ref{s:overview:invariant}).

We implemented \sys in Python for Google's Agent Development Kit and evaluated
it on the AgentDojo~\cite{agentdojo} and Toolathlon~\cite{toolathlon} benchmarks
(\S\ref{s:eval}). Across four models, \sys stays within a few percent of an
unconstrained agent's task success rate. Even in the worst case of a fully
hijacked agent, \sys's policies rule out 93--95\% of attacks on AgentDojo and
62--85\% on Toolathlon (doubling the defense rate of tool allowlists for the latter).

This paper makes the following contributions:
\begin{enumerate} %
    \item A stateful policy language targeting autonomous AI agents performing long-horizon tasks (\S\ref{s:policy_lang});
    \item \sys, an automaton-based engine that synthesizes policies from trusted user prompts and enforces them over tool calls with actionable recovery feedback (\S\ref{s:system}); and
    \item An evaluation of \sys's effect on utility and \emph{worst-case} security, the latter using an analytical methodology that checks whether a policy admits \emph{any} attack trajectory by a fully hijacked agent (\S\ref{s:eval}).
\end{enumerate}

While prior systems~\cite{forge, agentspec, shieldagent} also express dependencies across tool calls, they rely on \emph{static} policies derived ahead of time (manually or with offline LLM assistance) from fixed domain rules (\S\ref{s:related}).\footnote{Progent~\cite{progent} generates task-specific policies with an LLM, but its policy language only constrains individual tool calls; to adapt across steps, it feeds tool outputs back to an LLM after every call to rewrite those rules (\S\ref{s:related}).}
This suits single-purpose agents, but not general-purpose agents whose appropriate actions may vary with every user request.
Beyond the design of \sys's language and enforcement engine, our work suggests that today's LLMs can synthesize expressive, fine-grained policies from natural language, providing strong protection across diverse tasks without humans writing or auditing policies.

\section{Background and Motivation}
\label{s:setup}

An \textbf{AI agent} pairs a large language model~(LLM) with \emph{tools}:
functions that gather information (\eg read an email) or cause side effects
(\eg execute a bank transfer). Users provide agents with 
natural-language \emph{task prompts} (\eg \emph{``execute the bank transfer that
I emailed Alice about''}); the agent carries out tasks through an
execution loop managed by a runtime control program (the \emph{harness}). At
each step of the loop, the harness supplies the LLM with its current context---%
\eg the task prompt and the history of tool calls and their return values---and
the LLM emits the next tool call to invoke or decides the task is complete.

This paper focuses on \emph{general-purpose} AI agents designed to handle a
wide range of tasks on a user's behalf. Examples include commercial personal
assistants such as Muse~\cite{muse} and Gemini Spark~\cite{geminispark}, as well
as open-source solutions like OpenClaw~\cite{openclaw} and Hermes~\cite{hermes}.
These agents are highly \emph{versatile}, aimed to handle ``everyday tasks to personal goals''~\cite{muse} with simple natural language prompts.

\subsection{Rogue actions}
\label{s:setup:rogue}
An AI agent can execute \emph{rogue actions}---inappropriate tool calls
diverging from the user's intent. These rogue actions can be triggered by
adversarial manipulation or arise spontaneously during benign operation. 
In adversarial settings, the primary threat vector is \emph{prompt injection} (PI).
An attacker who controls content the agent ingests---a web page, an email, a calendar invite---can embed instructions that hijack the LLM's trajectory.
State-of-the-art safeguards remain vulnerable~\cite{attackermovessecond,
gptred}, leaving PI a persistent source of rogue actions.
Rogue actions can also occur without any adversary. During long-horizon tasks,
an agent may drift from its objective, misinterpret tool outputs, or improvise
unintended workarounds. For instance, upon encountering ``not found'' during a
search, an overly persistent web-browsing agent scraped a site owner's contact
information and sent unsolicited emails~\cite{meltdown}; and a ``misaligned''
agent performed inappropriate actions to fulfill a benign
goal~\cite{openai_alignment, huggingface_openai}.

\subsection{The need for contextual policies}
\label{s:setup:contextual}

Keeping a program's behaviors within intended bounds is a classic problem in
systems: like an AI agent, a traditional program can be hijacked~\cite{alephone}
or have bugs that cause damage~\cite{steamrm}.
A standard defense interposes a reference monitor that enforces an explicit
\emph{policy} over the program's actions~\cite{saltzerschroeder,
enforceablepolicies}, such as syscall filters~\cite{janus, systrace, seccomp} or
path confinement~\cite{apparmor}.
Such \emph{static} policies are derived from the program's fixed
purpose---traditionally known at development time.

Unfortunately, static policies fail to bound general-purpose AI agents due to
these agents' versatility. Because they are meant to handle diverse tasks, the
appropriateness of a given tool call depends on the task at hand~\cite{conseca}:
\code{send_money} is appropriate when \emph{``refunding the person who sent me
money last week''}, but is theft when \emph{``summarizing my inbox.''} For a
general-purpose agent, the only purpose fixed in advance---to do what the user
asks---is too broad to yield a meaningful static policy: it must either permit
\code{send_money} (and allow theft) or forbid it (and prevent appropriate
refunds).

To resolve this tension, we adopt the framework of contextual agent
security~\cite{conseca}: rather than fixing a policy in advance, synthesize a
\emph{contextual} policy just in time from the user's task prompt,
and enforce it over the agent's tool calls. Because the task prompt
is written in natural language, translating it into a precise policy requires an
LLM. While relying on a (policy generator) LLM to constrain an (agent) LLM may
seem circular, the policy generator is much less susceptible to sources of
rogue actions~(\S\ref{s:setup:rogue}).
First, policy generation is isolated from untrusted content, operating solely on
the task prompt and tool documentation, so an attacker has no channel to subvert
policy generation.%
\footnote{This mirrors a traditional program that determines which privileges it
needs from its (trusted) command-line arguments, drops the rest, and only then
begins to process untrusted input such as network packets~\cite{capsicum}.}
Second, translating a prompt into a policy is a well-scoped specification task
rather than an open-ended execution loop, avoiding the multi-step drift and
improvisation that can cause benign failures.

\subsection{The need for a stateful policy language}
\label{s:setup:good}

\begin{figure}[t]
\centering
\footnotesize
\begin{tikzpicture}[
  scale=1.25,
  lbl/.style={font=\scriptsize, inner sep=1pt},
]
\draw[draw=black!35, rounded corners=2pt]
  (0,0) rectangle (5.6,2.34);
\node[lbl, black!55, anchor=north west] at (0.1,2.30)
  {all trajectories};
\draw[draw=black!55, fill=black!7]
  (0.35,0.18) rectangle (5.25,1.98);
\node[lbl, black!65, anchor=north west] at (0.45,1.93)
  {stateless policy};
\draw[draw=green!40!black, fill=green!12, rounded corners=4pt]
  (0.9,0.43) -- (3.2,0.36) -- (4.35,0.60) -- (4.4,1.20)
  -- (3.7,1.63) -- (1.6,1.67) -- (0.85,1.29) -- cycle;
\node[lbl, green!35!black, anchor=north west] at (1.52,1.57)
  {stateful policy};
\draw[draw=blue!50!black, fill=blue!20, line width=0.4pt]
  plot[smooth cycle, tension=0.75] coordinates {
    (1.45,0.75) (2.05,0.60) (2.5,0.83) (3.15,0.65)
    (3.42,0.91) (3.0,1.03) (3.3,1.19) (2.55,1.12)
    (2.1,1.25) (1.8,1.01) (1.35,0.96) };
\node[lbl, blue!35!black] at (2.4,0.89) {appropriate trajectories};
\end{tikzpicture}
\caption{Policy tightness depends on language expressiveness. If a policy admits trajectories beyond the task's \emph{appropriate trajectories}, it also permits rogue actions. A \emph{stateless policy} (\eg an allowlist) over-approximates because it cannot reference changing runtime context. A \emph{stateful policy} more tightly encloses appropriate behavior.}
\Description{Policy tightness depends on language expressiveness.}
\label{fig:tightness}
\end{figure}
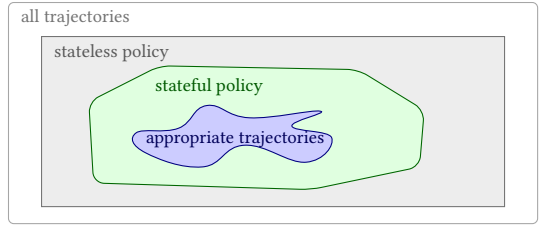

If we think of policy as a predicate over \emph{trajectories}---the sequence of
tool calls and return values---an ideal policy would admit every
\emph{appropriate trajectory} that an agent could take to fulfill the user's
task---and deny all others.
How closely a policy can approximate this ideal is ultimately bounded by the
expressiveness of the policy language (Figure~\ref{fig:tightness}). Consider
again \emph{``refund the person who sent me money last week.''} In a
coarse policy language (\eg a tool allowlist), the tightest policy that
admits all appropriate trajectories restricts the agent to calling
\code{list_transactions} and \code{send_money}. While this blocks irrelevant
tools like \code{send_email}, it permits sending any amount to any account,
leaving the core risk unaddressed. Even adding argument constraints does not
suffice: the agent only discovers the appropriate recipient and refund amount
after it reads the transaction history at runtime.

An effective agent policy language must thus be \emph{stateful}: what an agent
is permitted to do next must depend not only on the task prompt, but also on its
prior actions and observed data. A stateful policy can express exact constraints
for this example: the agent must first \code{list_transactions}, and then
\code{send_money} only to a sender and for an amount appearing in the past
week's transactions. \sys contributes such a stateful policy language, together
with a runtime engine that generates and enforces policies as the agent
executes.

Even in an expressive, stateful policy language like \sys's, there is generally
no policy that admits \emph{only} appropriate trajectories (hence the remaining
gap in Figure~\ref{fig:tightness}): defining appropriateness can depend on
natural-language semantics that resist exact specification, such as whether an
email body faithfully summarizes a document. (\sys addresses this gap in a
best-effort manner via scoped AI judges---see \S\ref{s:lang:judge}.)
Nevertheless, statefulness allows for much tighter policies than stateless
alternatives and is essential to blocking the worst outcomes in practical tasks.

\section{System Overview}
\label{s:overview}

\begin{figure}[t]
\centering
\footnotesize
\begin{tikzpicture}[
  >={Stealth[length=4pt]},
  box/.style={draw, rounded corners=1.5pt, align=center, inner sep=3pt,
              minimum height=6.5mm, font=\footnotesize},
  sys/.style={box, fill=green!8, draw=green!35!black},
  ext/.style={box, fill=black!5},
  lbl/.style={font=\tiny, inner sep=1.5pt},
]
\node[ext] (user)  at (0.0, 1.35) {Task\\prompt};
\node[sys] (gen)   at (2.25,1.35) {Policy\\generator};
\node[sys] (pol)   at (4.35,1.35) {Policy};

\node[ext] (agent) at (0.0, 0.0)  {Agent};
\node[sys] (enf)   at (4.35,0.0)  {Enforcer};
\node[ext] (tools) at (6.35,0.0)  {Tools};

\node[draw=green!35!black, dashed, rounded corners=2pt,
      fit=(gen)(pol)(enf), inner sep=3mm] (sysbox) {};
\node[lbl, green!35!black, anchor=south west]
      at ([xshift=1mm]sysbox.north west) {\sys};

\draw[->] (user) -- (gen);
\draw[->] (user) -- (agent);
\draw[->] (gen) -- (pol);
\draw[->] (pol) -- (enf) node[lbl, midway, right] {compile};

\draw[->] ([yshift=1.7mm]agent.east) -- ([yshift=1.7mm]enf.west)
      node[lbl, midway, above] {proposed tool call};
\draw[->] ([yshift=-1.7mm]enf.west) -- ([yshift=-1.7mm]agent.east)
      node[lbl, midway, below] {deny + reason};

\draw[->] ([yshift=1.7mm]enf.east) -- ([yshift=1.7mm]tools.west)
      node[lbl, midway, above] {allow};
\draw[->] ([yshift=-1.7mm]tools.west) -- ([yshift=-1.7mm]enf.east)
      node[lbl, midway, below] {result};
\end{tikzpicture}
\caption{\sys architecture. Before execution begins, the \emph{policy generator} translates the task prompt into a task-specific policy, reading no external content. During execution, the \emph{enforcer} checks every proposed tool call against the policy, returning feedback on denials and recording tool results for use in later checks.}
\Description{\sys architecture.}
\label{fig:arch}
\end{figure}
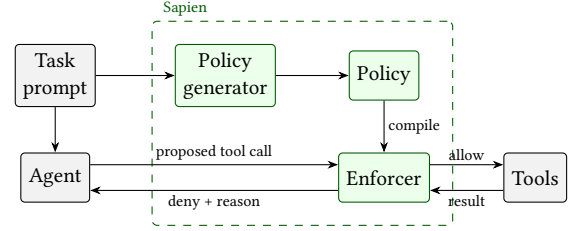
\sys aims to prevent rogue actions by automatically denying tool calls that
clearly fall outside the user's intent. To achieve this, \sys interposes on the
agent's harness to mediate between the agent and its tools.

\sys operates in two phases: ahead-of-time policy generation and runtime enforcement (Figure~\ref{fig:arch}).
\emph{Generation} runs once per task before the agent begins work: given the
task prompt and tool documentation, \sys invokes a generator LLM that
synthesizes a policy scoped to that task. Each policy governs a single task.
\emph{Enforcement} runs on every tool call the agent proposes.
The enforcer checks the call against the policy, executing it only if the policy
admits the call. Rather than escalating blocked calls to the user---which shifts
the burden of enforcement onto the human---\sys returns the denial to the agent
so the agent can correct itself.

\subsection{Threat Model and Assumptions}
\label{s:threat}

We assume an adversary who can compromise the agent LLM---whether indirectly via
prompt injections in tool outputs, or directly via memory poisoning or modifying
its weights---and thereby force it to propose arbitrary tool calls.
By contrast, we assume the policy-generator LLM is properly trained and
uncompromised: it faithfully follows \sys's instructions so long as its inputs
contain no malicious data. We \emph{do not} assume that the generator
can resist prompt injection; this is why \sys isolates it from untrusted
content.

We make three further assumptions about the agent's environment. First, the
user's session is uncompromised, so the task prompt genuinely originates from
the user. Second, the tools available to the agent behave as specified by their
documentation, which grounds policy generation. Third, the harness and \sys's
interposition cannot be bypassed and mediate every proposed tool call.

\subsection{The Policy Integrity ``Invariant''}
\label{s:overview:invariant}

Policy generation is the stage at which user intent enters the system, and \sys
rests on an (informal) policy integrity ``invariant'': \emph{the generated
policy reflects user intent}. The requirement is what makes enforcement
meaningful: an adversary-controlled policy is equivalent to no policy at all.
Because LLMs cannot reliably resist prompt injection, \sys guards this invariant
by design: the generator LLM sees only the user's task prompt and tool
documentation.
With adversarial inputs excluded, upholding the invariant only requires 
the policy LLM to accurately translate user intent.

Crucially, this isolation applies to policy \emph{generation}, not policy
\emph{enforcement}. A stateful policy inherently decides whether to admit a
tool call based on data the agent has observed during execution---\eg verifying
that a \code{send_money} call matches a prior transaction record---even though
that data may contain adversarial content. Evaluating such data during
enforcement \emph{does not} violate our invariant, because the data is merely
input to a fixed policy check: even if an adversary can inject data at runtime,
they cannot alter the criteria under which actions are judged, leaving the rules
that allow or deny actions anchored to the user's intent.

\subsection{Non-Goals}
\label{s:overview:nongoals}

\textbf{\sys does not vet or second-guess the task prompt.}
\sys aims to align the agent's actions and the user's task prompt; it does not judge whether the prompt itself is benign or accurately states the user's thoughts.
If a user asks for something harmful (\eg drafting a scam email), blocking that request falls to safety alignment and content moderation, which are orthogonal to \sys.
If a user mis-states their intent---\eg asking to delete the wrong file---then the deletion falls within the scope of the task prompt, and \sys has no reason to deny it.
Finally, if a user provides a vague or ambiguous prompt \sys will generate a wide policy that might permit harmful calls. 
Catching user errors requires a separate mechanism that asks the user to double-check anomalous or high-risk actions even when they are in scope.

Furthermore, whether an in-scope action---such as executing a large wire transfer or deleting a repository at the user's request---warrants human confirmation depends on its irreversibility and the user's risk tolerance, factors that cannot be inferred from the task prompt alone.
Deployments requiring human-in-the-loop approval for high-stakes operations can place a confirmation policy downstream of \sys.

Because \sys mediates tool calls, it provides safety (preventing unauthorized
tool calls) but \textbf{not liveness}: a prompt injection that convinces the agent to
stop working or refuse the task produces no unauthorized tool call, and
\sys does not intervene.
Finally, a compromised agent could still \textbf{leak information through side channels}, \eg timing channels or the ordering of permitted calls because a policy regulates only which tool calls are permitted, but not when they occur or which path the agent takes among allowed options.

\section{The \sys Policy Language}
\label{s:policy_lang}

A \sys policy specifies allowable sequences of tool calls that an agent may execute to complete a task.
At runtime, a tool call is permitted only if the extended trajectory---the historical tool-call sequence appended with the candidate call---is a valid prefix of a sequence admitted by the policy.

To express these temporal and data constraints, \sys builds on regular
expressions---a compact, widely adopted formalism in runtime
verification~\cite{rvlogics,tracematches,traceslicing} that has also been used
to express privacy policies in prior work~\cite{ancile}. Because tool calls are
not bare symbols but carry structured arguments and return values, we draw on a
variant of \emph{regular expressions for data words}~\cite{libkin2015regular},
which allows policies to bind, carry, and compare data values across execution
steps.

Figure~\ref{fig:policy_grammar} gives the grammar of the \sys policy language,
which combines four primitives:
\begin{enumerate} %
    \item Sequential ordering (\S\ref{s:lang:base}): regular operators augmented with ``shuffle''~\cite{DBLP:journals/iandc/MayerS94,broda2018automata} to describe interleaving;
    \item Variable binding and predicates (\S\ref{s:lang:bind}): state-tracking mechanisms to check arguments against runtime data;
    \item Deferred holes (\S\ref{s:lang:defer}): placeholders that expand into generated sub-policies during execution; and
    \item Semantic predicates (\S\ref{s:lang:judge}): natural-language guards delegated to a scoped LLM classifier.
\end{enumerate}
We next explain these primitives using a running example of an assistant agent.

\begin{figure}[t]
\centering
\footnotesize
$\begin{array}{@{}r@{\;\;}c@{\;\;}l@{\qquad}l@{}}
e & ::= & 1 & \text{empty trace} \\
  & \mid & c \;\mid\; c@x & \text{call pattern, may bind to } x \\
  & \mid & e_1 \cdot e_2 & \text{sequence: } e_1\text{, then } e_2 \\
  & \mid & e_1 + e_2 & \text{choice: either } e_1 \text{ or } e_2 \\
  & \mid & e^* \;\mid\; e^+ & \text{zero or more; one or more} \\
  & \mid & e_1 \shuffle e_2 & \text{both, interleaved in any order} \\
  & \mid & \mathsf{defer}(g) & \text{hole filled at run time} \\[3pt]
c & ::= & a \;\mid\; a(k{=}v, \dots) & \text{tool name, argument constraints} \\[3pt]
v & ::= & \textit{literal} \;\mid\; \mathsf{PREDICATE}(p) & \text{value constraint}
\end{array}$
\caption{The \sys policy language. $a$ is a tool name, $x$ a variable, $p$ a
named predicate, and $g$ a named generator; $p$ and $g$ are Python
functions supplied alongside the expression. A call pattern
$a(k_1{=}v_1,\dots)$ matches a call to $a$ whose argument $k_i$ satisfies
$v_i$; arguments not mentioned are unconstrained.
}
\Description{The \sys policy language.}
\label{fig:policy_grammar}
\end{figure}

\subsection{Regular operators}
\label{s:lang:base}

The classical regular operators apply directly to tool calls:

\begin{lstlisting}
read_inbox . send_email    # first read, then send
read_inbox + read_calendar # one or the other
search_web*                # any number of times
\end{lstlisting}

\noindent
Sequencing alone introduces stateful enforcement: under \code{read_inbox . send_email}, a \code{send_email} call is denied initially, but permitted after the inbox has been read.

To support tasks composed of independent subtasks, the language additionally includes the \emph{shuffle} operator~\cite{DBLP:journals/iandc/MayerS94,broda2018automata}:

\begin{lstlisting}
# both subtasks; steps may interleave:
(read_inbox . send_email) || (search_web . post_msg)
\end{lstlisting}

\noindent
This policy permits tool calls from both subtasks to interleave in any order, while ensuring that each subtask's internal ordering is preserved (\eg \code{send_email} only after \code{read_inbox}).
Shuffle addresses a common pattern in agent workflows: independent subtasks that need not follow a fixed order.\footnote{Expressing such flexibility using choice (\eg \code{a.b + b.a}) would incur combinatorial explosion as subtasks multiply; shuffle instead captures all step-by-step interleavings in a single, compact expression.}

\subsection{Binding and predicates}
\label{s:lang:bind}

While regular operators govern the order of tool calls, call patterns govern tool calls' arguments.
Each argument in a pattern matches against either a fixed literal or a \emph{predicate} (\code{PREDICATE(p)}), where \code{p} takes the proposed argument value and returns a boolean.
To express data dependencies across steps, a pattern may bind the call's arguments and return value to a \emph{variable} using \code{@x}; \code{x} can be used as a Python variable for subsequent predicates.
Unlike formalisms designed for static analysis, which restrict data guards to simple equality checks for decidability~\cite{libkin2015regular}, \sys evaluates predicates only at runtime against concrete arguments and variable bindings.
Predicates can therefore be written as Python functions, allowing the policy LLM to synthesize both the policy expression and predicate logic using its coding ability.

Consider a policy for the prompt \emph{``Pay my Acme invoice''}:

\begin{lstlisting}
list_invoices@invoices
  . pay_invoice(vendor="Acme",
                amount=PREDICATE(matches_invoice))

def matches_invoice(amt):
  return any(inv.vendor == "Acme"
             and inv.amount == amt
             for inv in invoices.result)
\end{lstlisting}

\noindent
This policy blocks three failures:
(1)~paying before reading the invoice list---blocked by \emph{call ordering};
(2)~paying a different vendor---blocked by the \emph{literal constraint}; and
(3)~paying a fabricated amount---blocked by the data-dependent \emph{predicate}.

A binding under iteration (\code{*} or \code{+}) accumulates a \emph{list} of records in execution order: \eg \code{(read_email}\code{@emails)*} saves \code{read_email} results in \code{emails}, each with its own \code{args} and \code{result}.
A predicate can then check a proposed argument against what the agent actually observed:

\begin{lstlisting}
list_emails . (read_email@emails)*
  . forward_email(to="accounting@corp.com",
                  email_id=PREDICATE(is_acme_invoice))

def is_acme_invoice(email_id):
  return any(e.args.email_id == email_id
             and e.result.sender == "billing@acme.com"
             for e in emails)
\end{lstlisting}

\noindent
No matter how many messages the agent reads (and in what order), it may forward only one from Acme it actually read.

\subsection{Deferred policy generation}
\label{s:lang:defer}

Some tasks reveal what the agent must do only as it executes.
Consider the prompt \emph{``Do the tasks on my TODO list at todo.io/bob''}.
While the agent's first step should clearly fetch that URL, any following tool calls depend entirely on the content of the page.
With the language features introduced so far, \sys would have no choice but to
allow arbitrary tool calls after the fetch---which \emph{is} the tightest
policy that admits all appropriate trajectories, since any tool may be
appropriate for whatever tasks the page lists. 

\sys addresses this limitation by introducing the \code{defer} operator, which places a \emph{hole} in the policy to be expanded at runtime once relevant data is available:

\begin{lstlisting}
get_page(url="todo.io/bob")@todo . defer(gen_todo)

def gen_todo():
  return f"Do these tasks: {todo.result}"
\end{lstlisting}

\noindent
Syntactically, \code{defer(g)} names a \emph{generator} \code{g}: a Python function supplied alongside the policy expression, which takes no arguments and returns a policy-generation instruction string.
At runtime, a defer-hole does not match a tool call; encountering \code{defer}
obliges the agent to call a \emph{resolve} action naming the generator,
at which point \sys calls \code{g}. This reads the bound variables in scope
(here, \code{todo.result}), composes the sub-task instruction, invokes the
policy generator on it (\S\ref{s:sys:gen}), and splices the generated sub-policy
into the hole (retaining any policy following the hole). The agent then resumes
under the expanded policy.

\textbf{\emph{Deferral and the policy integrity invariant.}}
Deriving a sub-policy from a fetched webpage might seem at odds with isolating
policy generation (\S\ref{s:overview:invariant}), which kept runtime data out of
the policy generator. But recall that isolation exists only to uphold the policy
integrity invariant: that the enforced policy reflects user intent. When the
user asks the agent to carry out tasks listed on a webpage, they delegate the
specification of those tasks to the page \emph{as if} the page had
been written directly into the task prompt. If the page contains prompt
injection, the sub-policy will be wrong for the same reason a tampered task
prompt yields the wrong initial policy. (Indeed, an attacker who can edit
the user's TODO list does not need to prompt inject the model, and can simply write malicious tasks directly!) Using \code{defer} thus helps to uphold the
invariant: a policy that expands to admit only the
tasks on the page reflects user intent more closely than one that permits
arbitrary actions after the fetch.

\subsection{Semantic predicates}
\label{s:lang:judge}

Not every constraint can be expressed as a deterministic check over structured data.
Consider the task prompt \emph{``Summarize \code{notes.md} and email it to
Alice.''} Verifying that the outgoing email faithfully summarizes the document
requires semantic judgment over unstructured text. For such cases, a predicate
may call \code{verify_with_ai}, a helper provided by the \sys runtime that poses
a yes/no question to an LLM:

\begin{lstlisting}
read_doc(name="notes.md")@notes
  . send_email(to="alice@corp.com",
               body=PREDICATE(is_summary))

def is_summary(body):
  return verify_with_ai(
    f"Is '{body}' a faithful summary of "
    f"'{notes.result}', with nothing extra?")
\end{lstlisting}

\noindent
Such scoped checks differ from using an LLM judge to vet agent behavior wholesale~\cite{\llmjudge} in two key ways.
First, a check is narrow and local; here, it compares two specific strings rather than asking an open-ended question about whether an action is intended or safe.
Second, the judge only receives inputs needed to answer that question (here, the source document and the proposed email), rather than the agent's full execution trace, reducing exposure to prompt injection.

The judge may still fall for prompt injection because it may inspect untrusted
text. However, a semantic predicate can only \emph{narrow} the actions permitted
by a policy: even if a judge is tricked into returning \code{True}, the
surrounding deterministic policy (reading \code{notes.md} and sending a single
email to \code{alice@corp.com}) remains strictly enforced. Semantic predicates
thus enable enforcing additional semantic properties that cannot be
checked in Python code alone.

\subsection{End-to-end example}
\label{s:lang:together}

We now put the four primitives together on the TODO-list task
(\S\ref{s:lang:defer}), \emph{``Do the tasks on my TODO list at todo.io/bob''}.
From the user's prompt, the policy generator produces an initial policy that
permits a single call, fetching \mbox{\code{todo.io/bob}}:

\begin{lstlisting}
get_page(url="todo.io/bob")@todo . defer(gen_todo)

def gen_todo():
  return f"Do these tasks: {todo.result}"
\end{lstlisting}

\noindent
Suppose the fetched page lists two tasks: forwarding Acme's invoice to
accounting (\S\ref{s:lang:bind}) and emailing Alice a summary of
\code{notes.md} (\S\ref{s:lang:judge}). Once \code{get_page} returns and \sys
binds the response to \code{todo}, the agent calls \emph{resolve} for
\mbox{\code{defer(gen_todo)}}. \sys then evaluates \mbox{\code{gen_todo()}}
using its stored \mbox{\code{todo}}, invokes the policy generator on the
resulting instruction, and splices the following sub-policy into the hole:

\begin{lstlisting}[escapeinside={(*@}{@*)}]
( list_emails . (read_email@emails)*
  . forward_email(
      to="accounting@corp.com",
      email_id=PREDICATE(is_acme_invoice)) )
|| ( read_doc(name="notes.md")@notes
     . send_email(
         to="alice@corp.com",
         body=PREDICATE(is_summary)) )

def is_acme_invoice(email_id): ...  # (see (*@\S\ref{s:lang:bind}@*))
def is_summary(body): ...           # (see (*@\S\ref{s:lang:judge}@*))
\end{lstlisting}

\noindent
The spliced sub-policy joins the two subtasks with shuffle (\code{||}), allowing
the agent to interleave their individual steps 
while preserving each subtask's internal read-before-write ordering. 
The policy admits at most two mutating calls, \code{forward_email} and
\code{send_email}, with recipients fixed by literals and payloads constrained by
data-dependent and semantic predicates on bound variables.

\section{The \sys Policy Engine}
\label{s:system}

This section describes \sys's runtime engine: how it integrates with an agent harness (\S\ref{s:impl:integration}), how it compiles a policy into an automaton and enforces it across tool calls (\S\ref{s:sys:enforce}), and how our prototype is implemented (\S\ref{s:sys:proto}).

\subsection{Integrating with an Agent Harness}
\label{s:impl:integration}

\sys is agnostic to both the agent framework and the underlying LLM.
At the core of \sys is a stateful enforcer interface with three methods:
\begin{itemize}[nosep,leftmargin=*]
\item \code{init(u)}: generates and installs a policy from task prompt~$u$.
\item \code{can\_advance(a, p)} $\to (ok, reason)$: checks whether a proposed call to tool~$a$ with arguments~$\mathit{p}$ is permitted given the execution trace so far, without modifying enforcer state; if not, provides feedback.
\item \code{advance(r)}: records the completion of the allowed tool call with return value~$r$, advancing the enforcer state.
\end{itemize}

\noindent
Integrating \sys into an agent harness requires calling \code{init(u)} when a
user request arrives, calling \code{can\_advance(a, p)} before executing each
proposed tool call, and calling \code{advance(r)} once an allowed call
returns~$r$. Beyond these three hooks, \sys requires only that the harness
register \sys's defer-resolution tool (\S\ref{s:lang:defer}) and allow \sys to
append instructions to the agent's context.

\textbf{\emph{Exposing the policy to the agent.}}
After generating a policy---both in \code{init} and when expanding
\code{defer}---\sys appends the policy expression, the predicate definitions,
and the generator's rationale to the agent's context. Seeing the active policy
helps the agent plan tool calls that satisfy the constraints and informs it 
when it must resolve a \code{defer} hole.

\textbf{\emph{Denial feedback.}}
When \code{can_advance} rejects a tool call, it returns a diagnostic message
identifying which argument failed which predicate along with that predicate's
source code. 
This feedback may help the agent recover in two situations: when a generated
policy is imperfect and blocks some legitimate trajectories for the task while
permitting others, or when the agent is misled by a prompt injection and the
denial prompts it to resume the user's original task.

\subsection{Enforcing a Policy}
\label{s:sys:enforce}

This section gives a high-level overview of how the enforcer compiles a policy into a nondeterministic finite automaton
(NFA) and steps it as the run proceeds.
\sys uses a standard syntax-directed translation over the policy AST following Thompson's construction~\cite{thompson}, mapping each sub-expression to an NFA with a single entry state and a single exit state.
A base call pattern compiles to two states connected by a single transition
carrying the tool name, parameter constraints, and optional variable binding.
Composite operators combine the sub-automata of their operands: concatenation
($\cdot$), choice (\code{+}), and iteration (\code{*}, \code{+}) connect entry
and exit states with $\varepsilon$-transitions, while shuffle (\code{||}) takes
the Cartesian product of its two operands' state
sets~\cite{DBLP:journals/iandc/MayerS94,broda2018automata}. 

\textbf{\emph{Tracking configurations.}}
Unlike a classical NFA, where two paths at the same state admit the same future
actions, transitions in \sys bind completed call records to variables that later
predicates may depend on. Thus, two paths reaching the same state with different
variable bindings can permit different future calls.
To handle this, the enforcer tracks active \emph{configurations} $(q, \rho)$, each pairing an automaton state~$q$ with a valuation~$\rho$ of the variables bound along that path.
On \code{can\_advance(a, p)}, the enforcer checks if any active configuration can (after $\varepsilon$-transitions) take a transition for tool~$a$ whose parameter constraints hold under~$\rho$.
If no configuration matches, the call is denied.
Otherwise, once the call returns~$r$, \code{advance(r)} steps every matching
configuration to its target state, extends its valuation~$\rho$ if the
transition binds a variable, and discards non-matching configurations.

\begin{figure}[t]
\small
\captionsetup[subfigure]{aboveskip=2pt,belowskip=6pt}
\begin{subfigure}{\linewidth}
\begin{lstlisting}
( search_emails@inv
  . forward_email(to="accounting@corp.com",
                  email_id=PREDICATE(is_acme_invoice)))
|| ( search_emails@bob
     . reply_email(email_id=PREDICATE(is_bob_email),
                   body=PREDICATE(answers_bob)) )
\end{lstlisting}
\caption{The policy.}
\label{fig:bnfa:policy}
\end{subfigure}

\begin{subfigure}{\linewidth}
\centering
\begin{tikzpicture}[
  >={Stealth[length=3.5pt]},
  st/.style={rounded corners=2pt, draw=black!70, inner sep=1.6pt, font=\scriptsize},
  hl/.style={->, draw=darkred, line width=0.9pt},
  pl/.style={->, draw=black!35},
  el/.style={font=\scriptsize, inner sep=1.2pt},
  hel/.style={el, text=darkred}
]
  \node[st]  (S00) at (0,     0)    {$L_0,R_0$};
  \node[st]  (S10) at (1.72,  0)    {$L_1,R_0$};
  \node[st]  (S20) at (3.44,  0)    {$L_2,R_0$};
  \node[st]  (S01) at (0,    -0.85) {$L_0,R_1$};
  \node[st]  (S11) at (1.72, -0.85) {$L_1,R_1$};
  \node[st]  (S21) at (3.44, -0.85) {$L_2,R_1$};
  \node[st]  (S02) at (0,    -1.7)  {$L_0,R_2$};
  \node[st]  (S12) at (1.72, -1.7)  {$L_1,R_2$};
  \node[st]  (S22) at (3.44, -1.7)  {$L_2,R_2$};
  \draw[->] (-0.8, 0) -- (S00);

  \draw[hl] (S00) -- node[hel, above] {$\ell_1$} (S10);
  \draw[hl] (S01) -- node[hel, above] {$\ell_1$} (S11);
  \draw[pl] (S02) -- node[el,  above] {$\ell_1$} (S12);
  \draw[pl] (S10) -- node[el,  above] {$\ell_2$} (S20);
  \draw[hl] (S11) -- node[hel, above] {$\ell_2$} (S21);
  \draw[pl] (S12) -- node[el,  above] {$\ell_2$} (S22);

  \draw[hl] (S00) -- node[hel, right] {$r_1$} (S01);
  \draw[hl] (S10) -- node[hel, right] {$r_1$} (S11);
  \draw[pl] (S20) -- node[el,  right] {$r_1$} (S21);
  \draw[pl] (S01) -- node[el,  right] {$r_2$} (S02);
  \draw[pl] (S11) -- node[el,  right] {$r_2$} (S12);
  \draw[hl] (S21) -- node[hel, right] {$r_2$} (S22);

  \node[anchor=west, font=\scriptsize] at (4.25, -0.85) {%
    \begin{tabular}{@{}r@{\,}l@{}}
      $\ell_1$: & \texttt{search\_emails@inv} \\
      $\ell_2$: & \texttt{forward\_email(...)} \\
      $r_1$:    & \texttt{search\_emails@bob} \\
      $r_2$:    & \texttt{reply\_email(...)} \\
    \end{tabular}};
\end{tikzpicture}
\caption{
  Compiled automaton: state $(L_i, R_j)$ tracks progress in each
  shuffle branch. Highlighted edges are taken in (c); $\varepsilon$-transitions elided.
}
\Description{The compiled automaton.}
\label{fig:bnfa:automata}
\end{subfigure}

\begin{subfigure}{\linewidth}
\centering
\scriptsize
\setlength{\tabcolsep}{3pt}
\renewcommand{\arraystretch}{1.1}
\begin{tabular}{@{}c l l@{}}
\toprule
\textbf{\#} & \textbf{Proposed tool call} & \textbf{Active configurations afterwards} \\
\midrule
0 & \emph{(initial)} & $((L_0, R_0),\ \emptyset)$ \\
1 & $c_1 =$ {\scriptsize\texttt{search\_emails("Bob")}} & $((L_1, R_0),\ \{\mathtt{inv}{=}c_1\})$; \\
  &                                                     & $((L_0, R_1),\ \{\mathtt{bob}{=}c_1\})$ \\
2 & $c_2 =$ {\scriptsize\texttt{search\_emails("Acme")}} & $((L_1, R_1),\ \{\mathtt{inv}{=}c_1, \mathtt{bob}{=}c_2\})$; \\
  &                                                      & $((L_1, R_1),\ \{\mathtt{inv}{=}c_2, \mathtt{bob}{=}c_1\})$ \\
3 & {\scriptsize\texttt{forward\_email(email\_id=9, ...)}} & $((L_2, R_1),\ \{\mathtt{inv}{=}c_2, \mathtt{bob}{=}c_1\})$ \\
4 & {\scriptsize\texttt{reply\_email(email\_id=7, ...)}}   & $((L_2, R_2),\ \{\mathtt{inv}{=}c_2, \mathtt{bob}{=}c_1\})$ \\
\bottomrule
\end{tabular}
\caption{Configuration trace. Step~2 yields two configurations at $(L_1, R_1)$
with swapped bindings; step~3's \code{is\_acme\_invoice} discards one.}
\label{fig:bnfa:trace}
\end{subfigure}
\vspace{-12pt}
\caption{Enforcement of a policy with two shuffled subtasks.}
\vspace{-12pt}
\label{fig:bnfa}
\end{figure}

Figure~\ref{fig:bnfa} illustrates enforcement on a policy for two inbox subtasks: forwarding Acme's invoice to accounting, and replying to Bob's question.
Each branch searches the inbox (\code{search_emails}) and then performs its action (\code{forward_email} or \code{reply_email}).
Because the agent may phrase a search query in different ways (\eg \mbox{\code{"Acme"}} vs.\ \mbox{\code{"invoice"}}), the policy leaves \code{search_emails} unconstrained on its \code{query} parameter and checks the returned message in the predicates of \code{forward_email} and \code{reply_email}.
Starting from $((L_0, R_0), \emptyset)$, the enforcer processes the four calls of Figure~\ref{fig:bnfa:trace} as follows:
\begin{enumerate}[nosep,leftmargin=*]
\item \code{search\_emails("Bob")} ($c_1$) matches both $\ell_1$ and $r_1$, giving two configurations with $\mathtt{inv}=c_1$ and $\mathtt{bob}=c_1$.
\item \code{search\_emails("Acme")} ($c_2$) takes the other search transition in each configuration, so both reach $(L_1, R_1)$ with swapped valuations.
\item \code{forward\_email(email\_id=9, ...)} can only follow $\ell_2$, whose predicate \code{is\_acme\_invoice} holds in only one valuation; the other configuration is discarded.
\item \code{reply\_email(email\_id=7, ...)} follows $r_2$, whose predicates are checked against the surviving binding $\mathtt{bob}{=}c_1$.
\end{enumerate}

\noindent\textbf{\emph{Resolving a defer.}}
A \code{defer(g)} hole compiles to a single transition for the defer-resolution tool (\S\ref{s:lang:defer}).
When the agent takes this transition, each matching configuration evaluates \code{g()} under its own valuation to obtain an instruction, invokes the policy generator, and compiles the returned sub-policy into a sub-automaton.
It then enters this sub-automaton and, upon reaching its exit state, continues from the state after the hole.
Because each configuration expands the hole under its own valuation into a separate sub-automaton, configurations with different bindings do not interfere.

\subsection{Prototype Implementation}
\label{s:sys:proto}

We implemented a prototype of \sys as a Python library and integrated it with Google's Agent Development Kit (ADK), as well as the AgentDojo~\cite{agentdojo} and Toolathlon~\cite{toolathlon} benchmark eval harnesses (\S\ref{s:eval}). All model calls are via the Gemini Enterprise Agent Platform~\cite{vertexai}.

\textbf{\emph{Generating policies.}}
\label{s:sys:gen}
The policy generator invokes an off-the-shelf LLM with a fixed system prompt
that defines the policy language, gives worked examples,
and instructs it to prefer deterministic predicates,
reserving \code{verify_with_ai} for properties that code cannot check.
For each task, \sys supplies the user's prompt and
tool schemas, including return types so predicates can access output fields.
Generation takes two LLM calls: the model drafts a rationale and
candidate policy, and then a self-correction call critiques the draft against a checklist (\eg valid syntax, necessary read-only tools are permitted) and emits the final policy
and Python block. Generating a deferred sub-policy works the same way,
with the generator's instruction string replacing the user's prompt and the parent policy as added context.

\textbf{\emph{Evaluating predicates and generators.}}
Predicates and \code{defer} generators are Python functions that reference bound policy variables (\eg \code{inv}, \code{bob}) as free variables, yet each configuration $(q, \rho)$ has its own valuation~$\rho$.
To support this without recompiling functions per configuration, the enforcer \code{exec}s the policy's Python block once in a custom globals dictionary, where Python resolves free variables at call time.
To evaluate a function under $(q, \rho)$, the enforcer fills this dictionary with the call records in~$\rho$, calls the function, and then clears the bindings.

\textbf{\emph{Expanding holes.}}
Instead of the general per-configuration hole expansion from
\S\ref{s:sys:enforce}, our prototype uses a simplified AST-level substitution
that suffices for all tasks in our evaluation: it substitutes the generated
sub-policy into the policy AST, recompiles the automaton, and replays the
recorded call trace to reconstruct the active configurations. This
simplification requires (and enforces at runtime) that each hole is resolved at
most once and that its generator produces the same instruction string across all
configurations at the hole.

\section{Evaluation}
\label{s:eval}

Our evaluation measures the advantages of \sys's approach to balance utility-security tradeoffs and its practicality:
\begin{enumerate}[leftmargin=*]
    \item \textbf{Utility}: Does \sys impact the agent's task success rate? 
    \item \textbf{Security}: Can attackers achieve their goals (caused \eg by prompt injections) in the presence of \sys policies?
    \item \textbf{Efficiency}: What latency and overheads does \sys add to agent execution, and why?
\end{enumerate}

\vspace{10pt}

\textbf{\emph{Baselines.}}
We compare three agents that differ in what governs their tool calls: no
policy (``NoPolicy''), a \sys{} policy, and a per-task allow-list of tool names
generated from the task prompt in the style of Conseca~\cite{conseca}
(``AllowList''). AllowList stands in for coarse-grained, stateless contextual
policies: it allows or denies each tool for the whole task, regardless of
arguments or execution state, and is generated in a single LLM call (without
\sys's self-correction pass, \S\ref{s:sys:gen}).

\textbf{\emph{Benchmarks.}}
We evaluate \sys on AgentDojo~\cite{agentdojo} and Toolathlon~\cite{toolathlon}.
AgentDojo comprises 97~short tasks (most within five turns) set in synthetic environments across four suites---workspace, Slack, travel, and banking---each exposing a small set of 11--28 tools.
Toolathlon features 108~longer-horizon tasks (dozens of turns) situated in a realistic environment with a local file/shell workspace and 32~software applications---including GitHub, Notion, and Google Workspace---that together expose over 600~tools via MCP.

To prevent Toolathlon from granting agents blanket, unlimited capabilities via a
handful of ``almighty tools'' like \code{local_python_execute}, all agents
upfront deny access to \code{local_python_execute} (impacting a handful of tasks
that would otherwise use it for arbitrary computation), and we require \sys to
specify command predicates (\eg the command must be \code{grep}) for
\code{terminal_run_command}.

\textbf{\emph{Models.}}
We evaluate \sys using four models: Gemini~3.8 Flash, Gemini~3.1 Pro, Claude Sonnet~5, and Claude Opus~5.
Each run uses the same model for agent execution and policy generation (\sys or AllowList).
On AgentDojo, we run all 97~tasks (949~attack instances) on all four models.
For Toolathlon, we run all 108~tasks (212~attack instances) with Gemini~3.8 Flash and, due to cost, a random subset of 20~tasks (40~attack instances) with the other three models.

\textbf{\emph{Utility.}}
Utility is the fraction of tasks the agent completes in a benign environment,
measured using each benchmark's native completion criteria. AgentDojo compares
the final environment state or tool-call trajectory against a per-task
specification. Toolathlon runs a per-task verification script comparing the
agent's workspace against a ground-truth workspace and querying modified
external services.

\textbf{\emph{Analytical security.}}
In standard agent security evaluations, one plants prompt injections in tool
outputs and measures how often the agent executes them, conflating policy
enforcement with the model's built-in injection resistance. However, \sys is
designed as a second line of defense even when the model itself is compromised.
We therefore evaluate security \emph{analytically}: assuming an attacker has
already hijacked the agent, does the policy admit \emph{any} sequence of tool
calls that achieves the attacker's goal?

Concretely, each \emph{attack instance} pairs a user task and environment 
with an attacker goal and programmatic attack-success check. For AgentDojo, we
pair every user task in a suite with every ``injection task'' in that suite,
yielding 949 attack instances.%
\footnote{AgentDojo excludes instances whose user task never reads attacker-controlled data. We keep them because we assume the agent is already compromised, whether or not benign execution would encounter the injection.}
For Toolathlon, which defines no attacker goals, we craft 1--2 plausible goals per user task that target the task's own tools and data, yielding 212 attack instances.%
\footnote{For example, ``route student through dangerous construction zones'' on the task of computing a campus walking route to \code{route.txt}, or ``change candidate rejection into strong hire'' on the task of compiling interview feedback into \code{interview\_summary.md}.}

We then evaluate each attack instance under conservative assumptions, in which
only two kinds of LLM calls behave faithfully: \one{} initial policy generation, and
\two{} \code{defer} resolutions on inputs untainted by attacker-controlled data.
All other LLM calls are assumed to be compromised: the agent can propose
arbitrary tool calls, every \code{verify_with_ai} check passes, and any
\code{defer} hole resolved on instructions tainted by attacker-controlled data
permits all tool calls.

To decide attack success, we run three automated checks:
\one{} a \emph{required-tool check} concludes \emph{attack failure} if the policy completely forbids any tool necessary for the attack (annotated by us per goal);
\two{}~a \emph{replay check} concludes \emph{attack success} if the policy admits the benchmark's ground-truth attack trajectory (available in AgentDojo); and
\three{}~an \emph{adversarial search} concludes \emph{attack success} if an adversarial agent (Gemini~3.8 Flash), given the policy and goal, finds an exploit trajectory allowed by the policy.
We manually adjudicate any remaining instances
under the same assumptions.%

\subsection{Utility}
\label{s:eval:utility}

Any policy enforcement risks reducing utility: the policy generator may not anticipate every valid way to solve a task, so it may produce an overly tight policy. %

\begin{figure}
    \centering
    \includegraphics[width=\linewidth]{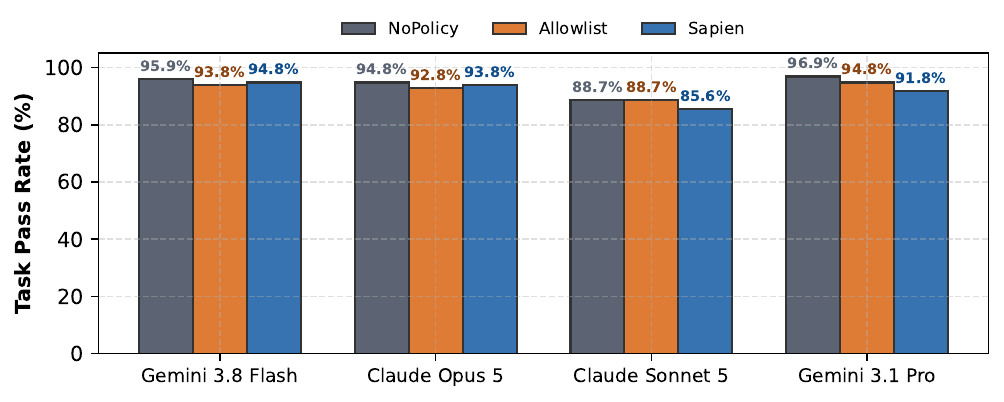}
    \caption{AgentDojo utility results across models. \sys achieves comparable task completion to NoPolicy and AllowList across all 97 tasks.}
    \Description{Bar chart showing task pass rates on the 97-task AgentDojo suite across four models for NoPolicy, AllowList, and \sys.}
    \label{fig:agentdojo:utility}
\end{figure}

\begin{figure*}
    \centering
    \includegraphics[width=\linewidth]{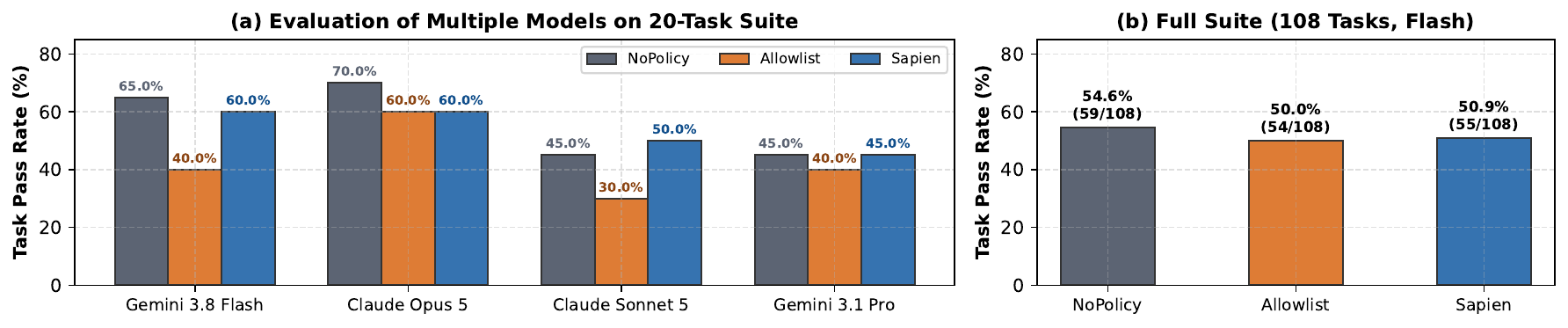}
    \caption{Toolathlon utility across models. \sys roughly matches NoPolicy and improves over generated AllowList policies.}
    \Description{Bar chart showing task pass rates on the 20-task multi-model suite and full 108-task suite of Toolathlon for NoPolicy, AllowList, and \sys.}
    \label{fig:tool:utility}
\end{figure*}

On AgentDojo (Figure~\ref{fig:agentdojo:utility}), this cost is small: across
all models, \sys completes 1--5 fewer of the 97 tasks than NoPolicy, and
AllowList completes 0--2 fewer. 
On Toolathlon (Figure~\ref{fig:tool:utility}), \sys comes within 4\% of
NoPolicy's task success rate on the full Toolathlon suite (with Gemini 3.8
Flash); on the 20-task subset across all four models, \sys succeeds on 43 of the
80 task runs, while NoPolicy succeeds on 45.

\sys fails at most 2 additional tasks on the Toolathlon 20-task benchmark
regardless of model.
 \sys achieves parity with Gemini 3.1 Pro; exceeds NoPolicy task success on
 Claude Sonnet, and fails an additional 1 and 2 tasks with Gemini 3.8 Flash and
 Claude Opus 5 respectively. AllowList performs worse than \sys and NoPolicy
regardless of model (with parity on Opus 5). 
\sys's improved utility with Sonnet occurs on a task in which NoPolicy reaches
the maximum number of turns; \sys's policy helps the agent
remain on the correct trajectory to solve the task (also see
\S\ref{s:eval:perf}).

\textbf{\emph{When \sys policies hurt utility.}}
\sys policies often prevent task completion because the policy generator does
not see the runtime environment, and makes incorrect assumptions about the data
or workflow. For example, its Python predicates may expect the wrong data format
or value (\eg matching \code{"London Luxury Hotel"} from the prompt when the
database stores \code{"London Luxury"}). Its \code{verify_with_ai} checks may
read the prompt too literally (\eg rejecting an update to a rent ``standing
order'' that the environment stores as a one-time payment). Finally, it may
restrict the workflow too tightly by omitting a needed read tool.

\textbf{\emph{Why \sys can have higher utility than AllowList.}}
One might expect a coarse policy language like AllowList to yield more permissive contextual policies (\S\ref{s:setup:good})---and thus higher utility---than \sys's fine-grained constraints.
But \sys has higher utility than AllowList on Toolathlon and on two of the four runs in AgentDojo, for two main reasons.

First, the AllowList generator must decide ``all-or-nothing'' upfront tool
permissions: allowing a sensitive tool (\eg \code{send_}\code{email}) permits it with
arbitrary arguments for the entire task, so the generator often withholds the
tool when the prompt does not explicitly require it.
\sys avoids this dilemma: argument predicates (\eg bounding file writes to
\code{/workspace/result.md}) and \code{defer} holes let the generator permit
sensitive tools while tightly bounding their use.%
\footnote{For example, on an AgentDojo task to carry out an emailed TODO list,
AllowList denied \code{send_email} (\emph{``The user did not
request sending any emails. Allowing \code{send_email} poses a severe risk of
\dots\ data exfiltration''}), causing the task to fail when the email asks for a
reply. \sys instead used \code{defer}, whose resolved sub-policy allowed
\code{send_email} only to the sender.}

Second, unlike a flat allow-list, \sys's staged policy structure gives the agent implicit guidance on the expected sequence of steps (consistent with \sys's lower average inferences per task on Toolathlon in \S\ref{s:eval:perf}).

\subsection{Security}

Under our analytical model, a hijacked agent can issue any sequence of tool
calls admitted by the policy, so NoPolicy blocks 0\% of attacks. AllowList
blocks an attack only when the attacker's goal requires a tool the user's task
does not need at all, and may permit attacks that abuse a required tool.
\sys closes this gap with fine-grained, stateful constraints.

\begin{figure}
\centering
\small
\begin{tabular}{lrrr}
\toprule
 & \multicolumn{3}{c}{\textbf{Analytical Defense Rate}} \\
\cmidrule(lr){2-4}
\textbf{Model} & \textbf{NoPolicy} & \textbf{AllowList} & \textbf{\sys} \\
\midrule
\textbf{Gemini 3.8 Flash} & 0\% & 86.4\% & \textbf{93.4\%} \\
\textbf{Claude Opus 5}    & 0\% & 85.8\% & \textbf{94.8\%} \\
\textbf{Claude Sonnet 5}  & 0\% & 86.8\% & \textbf{95.2\%} \\
\textbf{Gemini 3.1 Pro}   & 0\% & 82.7\% & \textbf{93.2\%} \\
\bottomrule
\end{tabular}
\caption{AgentDojo analytical security evaluation.}
\Description{Table showing the fraction of the 949 AgentDojo attack instances blocked by NoPolicy, AllowList, and \sys across four models.}
\label{tab:agentdojo_security}
\end{figure}

\begin{figure}
\centering
\small
\begin{tabular}{lrrr}
\toprule
 & \multicolumn{3}{c}{\textbf{Analytical Defense Rate}} \\
\cmidrule(lr){2-4}
\textbf{Benchmark} & \textbf{NoPolicy} & \textbf{AllowList} & \textbf{\sys} \\
\midrule
\multicolumn{4}{l}{\textit{20-Task Suite (40 Attack Instances / Model)}} \\
\textbf{Gemini 3.8 Flash} & 0\% & 30.0\% & \textbf{67.5\%} \\
\textbf{Claude Opus 5}    & 0\% & 22.5\% & \textbf{85.0\%} \\
\textbf{Claude Sonnet 5}  & 0\% & 30.0\% & \textbf{62.5\%} \\
\textbf{Gemini 3.1 Pro}   & 0\% & 35.0\% & \textbf{77.5\%} \\
\midrule
\textbf{Full 212 Attacks (Flash)} & 0\% & 32.4\% & \textbf{64.6\%} \\
\bottomrule
\end{tabular}
\caption{Toolathlon analytical security evaluation.}
\Description{Table showing the analytical defense rate of NoPolicy, AllowList, and \sys on the Toolathlon 20-task suite across four models and on the full 212-instance suite with Gemini 3.8 Flash.}
\label{tab:tool_security}
\end{figure}

\begin{figure*}[t]
\begin{minipage}[b]{\columnwidth}
    \centering
    \begin{subfigure}{\linewidth}
        \centering
        \includegraphics[width=.8\linewidth]{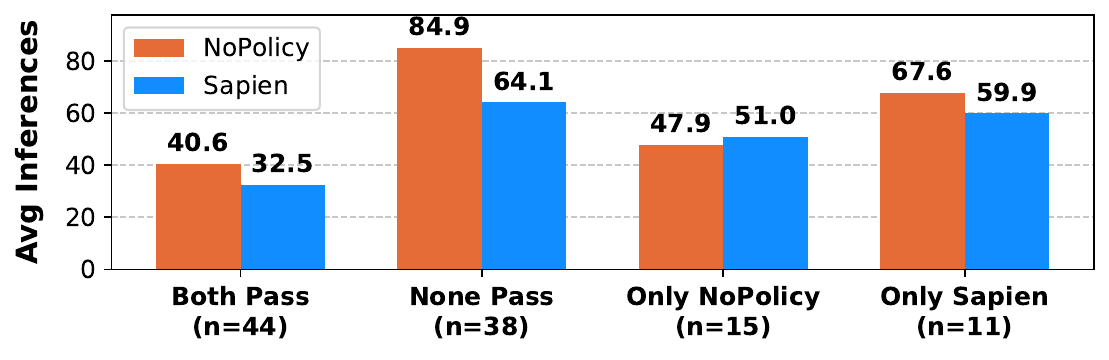}
        \caption{NoPolicy vs. \sys}
        \label{fig:eval:inferences:a}
    \end{subfigure}

    \begin{subfigure}{\linewidth}
        \centering
        \includegraphics[width=.8\linewidth]{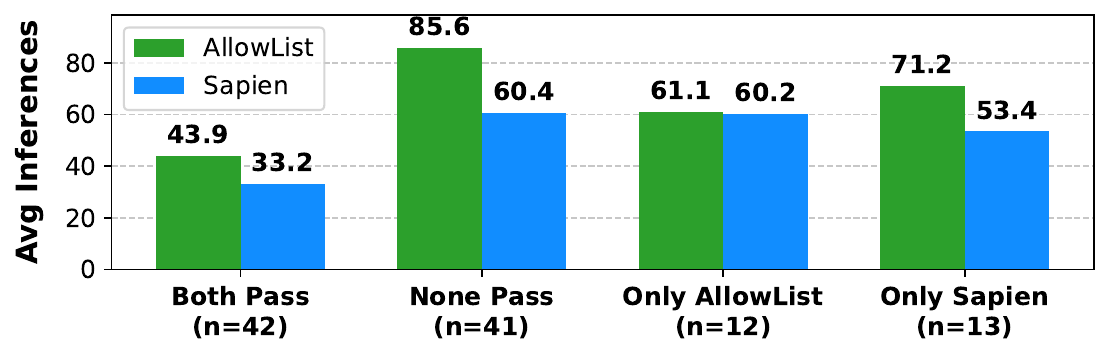}
        \caption{AllowList vs. \sys}
        \label{fig:eval:inferences:b}
    \end{subfigure}

    \caption{Average inferences per task by outcome (Full Toolathlon Suite, Gemini 3.8 Flash).}
    \Description{Two bar charts of average agent inferences per task on Toolathlon, comparing NoPolicy and AllowList with \sys across four outcome buckets.}
    \label{fig:eval:inferences}
\end{minipage}\hfill
\begin{minipage}[b]{\columnwidth}
    \centering
    \begin{subfigure}{\linewidth}
        \centering
        \includegraphics[width=.8\linewidth]{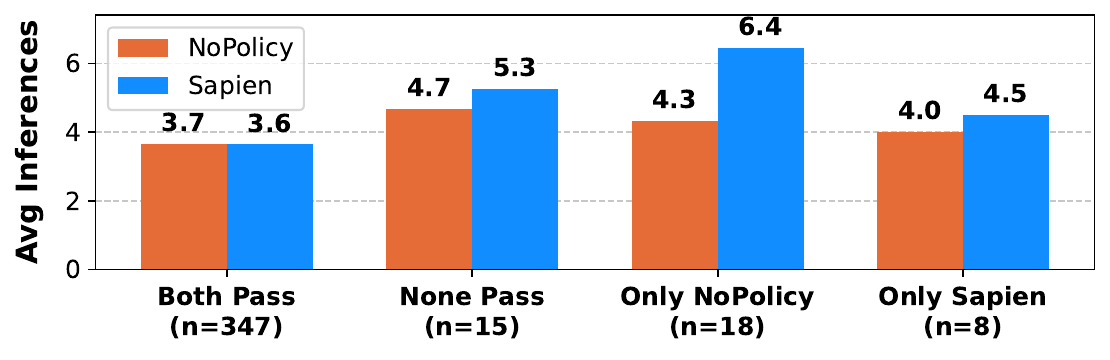}
        \caption{NoPolicy vs. \sys}
        \label{fig:eval:inferences:agentdojo:a}
    \end{subfigure}

    \begin{subfigure}{\linewidth}
        \centering
        \includegraphics[width=.8\linewidth]{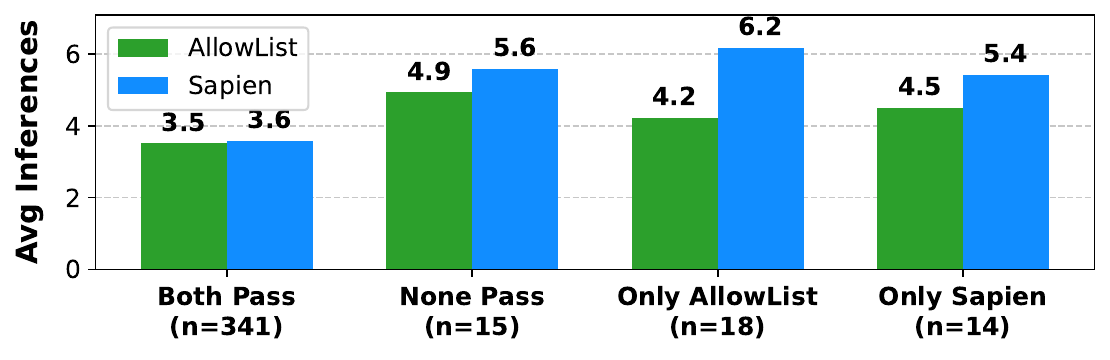}
        \caption{AllowList vs. \sys}
        \label{fig:eval:inferences:agentdojo:b}
    \end{subfigure}

    \caption{Average agent inferences per task by outcome (AgentDojo, 97 tasks pooled across the four models).}
    \Description{Two bar charts of average agent inferences per task on AgentDojo, comparing NoPolicy and AllowList with \sys across four outcome buckets.}
    \label{fig:eval:inferences:agentdojo}
\end{minipage}
\end{figure*}

On AgentDojo (Figure~\ref{tab:agentdojo_security}), \sys blocks 93.2--95.2\% of the 949 attack instances depending on the model, against 82.7--86.8\% for AllowList.
On Toolathlon (Figure~\ref{tab:tool_security}), where the attacker goals target tools the user task requires, \sys blocks 64.6\% of the 212 attack instances on the full suite with Gemini 3.8 Flash (against 32.4\% for AllowList), and 62.5--85.0\% of the 40 instances on the 20-task subset across the four models (against 22.5--35.0\% for AllowList).
Across both benchmarks, attacks that succeed under \sys fall into three categories:
\begin{enumerate}[leftmargin=*]
    \item \emph{Semantic attacks (82.1\% of unprevented attack instances on AgentDojo, 84\% on Toolathlon).}
    Here, distinguishing benign from malicious behavior is a semantic question that a deterministic predicate cannot decide.
    Usually the attack is carried by content the agent produces, such as file contents, message bodies, or the final response.%
    \footnote{A final response involves no tool call, so \sys cannot constrain it anyway. On AgentDojo, one attacker goal that only asks the agent to output a sentence accounts for 36\% of successful attack instances.}
    \sys either leaves this content unconstrained or checks it with \code{verify_with_ai}.
    In the remaining cases, the user's prompt delegates to untrusted data (\eg \emph{``follow the instructions in this notice''}), so the policy generator resolves a \code{defer} hole on attacker-controlled input.
    Because we assume the attacker can subvert any LLM call on untrusted input, we count both cases as allowing the attack.
    \item \emph{Imprecise policy (1.8\% of AgentDojo, 16\% of Toolathlon).}
    Here, \sys could have expressed a tighter policy but produced a looser one.
    On Toolathlon, often the policy pins only the GitHub repository or parent directory of a file write rather than the exact path, so an attacker can modify other files there.
    On AgentDojo, the cases come from a loosely matched URL, a payment update that pins which payment to edit but not the new recipient, and a misnamed parameter that left the real parameter unconstrained (checking policies against tool schemas would catch this).
    \item \emph{Permissive read access (16.1\% of AgentDojo).}
    To avoid blocking legitimate exploration, \sys often leaves read-only tools unconstrained when the prompt does not fully specify which resources to read.
    One AgentDojo attacker goal simply asks the agent to visit an attacker-chosen URL, which such policies allow.
    This is a tunable trade-off: constraining read tools would block the attack but could break tasks that need to browse unanticipated pages.
\end{enumerate}

\subsection{Efficiency}
\label{s:eval:perf}

Efficiency depends on the agent's inference count and latency, plus the policy-generation and policy-check overheads of \sys or AllowList.
The latencies we report are wall-clock times measured against production model-serving endpoints, which vary with capacity, load, and rate limits.
We report absolute numbers for completeness but focus on relative trends across configurations rather than exact numerical overheads.

\textbf{\emph{Inferences.}}
Figure~\ref{fig:eval:inferences} shows the average number of agent planning
inferences performed over all 108 Toolathlon tasks, split by which configurations succeeded.
\sys reduces the average number of turns by 19.9\% on tasks where both \sys and NoPolicy succeed; by 
24.5\% when both fail; and by 11.4\% on tasks where only \sys succeeds.
On tasks where the baseline succeeds but \sys fails, \sys has slightly higher inference counts (as expected: when blocked, the agent retries).
A similar pattern holds for \sys vs. AllowList, with \sys reducing inference
count even further than with NoPolicy. AllowList's high inference counts
(particularly with failed tasks) occur because if the rigid policy lacks a tool the
agent believes it requires, the agent retries until it hits a 200-retry limit.

On AgentDojo (Figure~\ref{fig:eval:inferences:agentdojo}), \sys does not reduce
inferences: most tasks finish in 3--4 turns, leaving little room for exploration.
As on Toolathlon, \sys requires more inferences when it fails and NoPolicy
succeeds (46--49\%) as the agent re-proposes blocked actions.
AllowList does not induce repeated retries on AgentDojo: the agent sees the
allow-list in its prompt and stops rather than calling an unlisted tool
(AllowList blocks only a single call across all models).

\textbf{\emph{Policy generation.}}
Policy generation runs once per task, so its share of total time is modest on Toolathlon's long-horizon tasks but dominant on AgentDojo's short ones.
With Gemini 3.8 Flash on the full Toolathlon suite, \sys spends on average \qty{148}{\second} per task generating its policy, compared to \qty{28}{\second} for AllowList.
On AgentDojo, \sys spends \qtyrange{43}{100}{\second} per task on policy generation depending on the model, compared to \qtyrange{18}{23}{\second} for AllowList.
\sys is slower than AllowList for two reasons: it synthesizes predicates and multi-phase automata rather than a flat tool list, and it runs a two-call self-correction pass (\S\ref{s:sys:gen}) where AllowList makes a single call. %
On both benchmarks, tasks with a deferred hole spend \qtyrange{74}{196}{\second} resolving it, again depending on the model.

\textbf{\emph{Policy checks.}}
When the agent executes an action, \sys and AllowList check this action against
the policy. 
These add $\approx$\qtyrange{15}{45}{\milli\second} of overhead
due to proxy interception and logging in Toolathlon; deterministic policy checks in both benchmarks 
take less than \qty{1}{\milli\second} per step.
Semantic predicates dominate when present: \code{verify_with_ai} calls (0.6 per AgentDojo task and 1.03 per Toolathlon task on average) add \qtyrange{1.4}{3.5}{\second} per AgentDojo task depending on the model.

Every blocked action induces an \emph{action reproposal} with policy feedback, \ie 
another agent inference, each of which adds on average \qtyrange{7.2}{10.7}{\second} on Toolathlon and \qtyrange{4.4}{8.5}{\second} on AgentDojo, depending on the model. On the full Toolathlon suite (Gemini 3.8 Flash), \sys blocks 2.16 actions per task on average (233 total), and AllowList blocks 0.75 (81 total). The agent resolves 84\% of \sys-blocked actions within one retry.
Blocks are far rarer on AgentDojo---only 0.14 per task on average (54 across the four models' 388 task runs)---but harder to recover from: only 31\% of denied calls (17 of 54) are followed by an allowed call, and 11 of the 20 tasks with a denial fail.
These blocks mostly fall on the task's mutating actions (\eg a predicate on the new amount of \code{update_scheduled_transaction} or on the body of \code{send_email}), which have no alternative route, so the agent either repeats the call or gives up.

\textbf{\emph{Overall performance.}}
On both benchmarks, \sys has the highest per-task latency.
On Toolathlon (Gemini 3.8 Flash, full suite), a task takes \qty{679}{\second} on
average with \sys, against \qty{577}{\second} with AllowList and
\qty{548}{\second} with NoPolicy. On AgentDojo, the mean task takes
\qtyrange{66}{144}{\second} with \sys across the four models, against
\qtyrange{30}{62}{\second} with AllowList and \qtyrange{12}{62}{\second} with
NoPolicy. 
The overhead comes from policy generation and policy checks, even
though \sys reduces the number of agent inferences on Toolathlon. Policy
generation accounts for 21\% of \sys's wall time on Toolathlon and 
65--69\% on AgentDojo (73--78\% including deferred resolution and
\code{verify_with_ai}).

\subsection{Takeaways}
\sys achieves utility parity with baselines across both benchmarks and all four
models, staying within a few tasks of an unconstrained agent. On long-horizon
tasks, \sys beats a tool allow-list because argument predicates and \code{defer}
holes let it permit sensitive tools under tight bounds rather than
deny them.
The same constraints deliver strong security against a hijacked agent. Under our
conservative analytical evaluation, \sys blocks over 90\% of attacks on
AgentDojo. On Toolathlon, where attacks abuse tools the task itself needs, it
blocks roughly twice as many as an allow-list. Most remaining attacks exploit
the \emph{semantics} of data the agent ingests or produces, which deterministic
predicates cannot judge.

The price we pay is latency, almost all from policy generation:
on short tasks, it can dominate runtime, but on longer tasks,
it is amortized over many steps and partly offset as \sys's
staged policies guide the agent to fewer planning inferences. \S\ref{s:disc}
discusses future directions to lower this overhead.

\section{Discussion}
\label{s:disc}

\sys policies face several limitations.
First, when an \textbf{under-specified user prompt} admits multiple valid
actions, \sys does not add precision and permits all of them. For example, a
request may be broad (\eg \emph{``handle my inbox''}) or one where multiple
targets match the user's description (\eg an inbox contains two different Bob
emails and the user asks to \emph{``reply to Bob''}). This allows unexpected
outcomes and adversarial steering (\eg replying to attacker-controlled ``Bob''
instead of the correct
Bob). 
As promising future work, \sys could detect such
under-specification~\cite{ambiguity} and prompt the user for
disambiguation before generating a policy.

The generator can also \textbf{get the policy wrong}. Because enforcement holds
relative to the generated policy, \sys ultimately relies on the
policy-generator LLM, which can err in either direction. A policy that is too
broad sacrifices security; a policy that is too tight reduces utility. While the
agent can often adjust after a denial, tasks can fail when an unexpected
obstacle (\eg a broken tool) requires a workaround the policy LLM did not
foresee. Nevertheless, our results and others'~\cite{progent, camel} demonstrate
that LLMs already can generate relatively accurate policies, and \sys will
transparently benefit from underlying model advances, whether through general
capability gains or targeted fine-tuning.

Furthermore, unlike deterministic checks, \textbf{LLMs evaluating semantic
predicates may be deceived by prompt injection}. However, the impact of fooling
a judge is bounded: an attacker can satisfy only that specific semantic
predicate, while the surrounding structural constraints---which tools may be
called, in what order, and with what arguments---remain deterministically
enforced. AI judges thus only further restrict actions where deterministic
checks fall short.

Finally,
\textbf{policy tightness depends on tool granularity}. For a focused tool like
\code{send_money(recipient, amount)}, the effect of a call is explicit in its
arguments and easy to constrain. By contrast, for a general-purpose execution
tool like \code{run_python(source)}, verifying whether the arbitrary source
string matches user intent requires program analysis beyond the reach of
argument predicates. Like other reference monitors, \sys is most effective when
agents are given narrow, well-scoped tools.
The tested benchmarks have no (AgentDojo) or few (Toolathlon) coarse-grained
tools, which we constrain appropriately in \sys (\S\ref{s:eval:utility});
a future direction can look into deriving granular tools from coarse ones.

\textbf{\emph{Improving latency and scalability.}}
To reduce \sys's latency overheads, which are mostly caused by generating policies and evaluating semantic predicates, \sys could better \emph{utilize caching}: prefix KV-caching of (static) tool schemas would eliminate redundant prefill cost, and \sys could reuse cached parameterized policy templates for similar tasks. %

\sys could also leverage \emph{model routing} based on complexity: routing
simple tasks and semantic checks to lightweight classifiers rather than frontier
models can reduce these inferences' latencies from seconds to tens of
milliseconds.

For large tool sets, retrieving only the \emph{top-k
task-relevant tools} via a lightweight semantic search prior to synthesis will
reduce both context length and reasoning complexity.
\sys policies also define exact set of transitions permitted at the
current configuration; thus, the harness could \emph{prune the agent input tool schemas}
to only currently allowed tools.

Finally, \sys could \emph{parallelize policy generation and agent execution}.
Agents could begin executing actions for a task as soon as \sys generates the
initial phase of the policy, so long as policy generation remains at least one stage ahead.

\section{Related Work}
\label{s:related}

\textbf{\emph{Policy-based defenses.}}
Prior works~\cite{progent, forge, conseca, camel, cellmate, nvidia, cloudflare,
openaiAgentsGuardrails, ace, saga, agentspec, provos2026ironcurtain} also defend
against agents compromised by prompt injection, model
poisoning, or malicious system prompts. Like \sys, they do so by checking agent actions against a (static or contextual) policy.
\sys contributes a stateful policy language whose fine-grained constraints can incorporate context discovered at runtime.

Works like Conseca~\cite{conseca}, CaMeL~\cite{camel}, and others~\cite{fsecure,
ace, shieldagent} use LLMs for policy generation and follow the dual-LLM
pattern~\cite{willisonisolation} to ensure the security monitor is itself not
compromised by untrusted inputs; \sys similarly isolates the policy generation
(\S\ref{s:overview:invariant}).
CaMeL~\cite{camel} policies are fixed control-flow graphs,
supporting only single-turn tasks where the policy itself must
execute the task, whereas \sys policies permit many trajectories.
CaMeL's IFC-based data disclosure policies are orthogonal and could complement \sys.

Progent's~\cite{progent} policies---predicated allow/deny rules with fallback
clauses and predefined modifications when certain conditions match---are either
static per agent, or generated from user instructions and regenerated after a
tool call if an LLM deems it necessary. \sys only ever makes policy decisions
off of trusted relevant context, and does not require an LLM call to check for
updates at every tool call.

FORGE~\cite{forge} supports manual, static Datalog policies constraining causal relationships across agent
executions (predicates on a DAG of actions). \sys policies are dynamic and can constrain action sequences of concurrent agents via shuffle;
\sys can be extended with causal predicates.

IsolateGPT~\cite{isolategpt} proposes an architecture to prevent
cross-LLM-application compromises by allowing only structured, user-approved
communication across app boundaries. \sys is orthogonal and can help generate
and enforce constraints both across and within applications.

\textbf{\emph{Defenses against compromised inputs.}}
IFC-based systems~\cite{fides, prudentia, rtbas, permissive} prevent untrusted inputs from influencing agent behavior but assume agents otherwise behave correctly and auto-approve any untainted action; they do not defend against
agents spontaneously taking malicious or erroneous actions.
These systems also require manually specified trust labels for data sources;
\sys does not rely on such labels, though they could help inform its predicates.

\textbf{\emph{Model-based security techniques.}}
Probabilistic defenses aim to use AI models to detect attacks~\cite{detect1, detect2, willisondetection, llamaguard}, or to distinguish untrusted from trusted inputs~\cite{spotlighting, yidefense, willisondelimiters, instructionhierarchy}. 
Others rely on an LLM to identify threats~\cite{\LLMguardrails}.
However, these remain susceptible to attacks introduced by untrusted
context~\cite{attackermovessecond, gptred}.
By contrast, \sys isolates policy generation from sources of prompt injection.

\section{Conclusion}
General-purpose agents need task-specific policies that track what the agent has done and learned.
\sys provides a stateful policy language and an engine that enforces such policies over tool calls.
Our evaluation shows that \sys preserves utility while tightly constraining even a fully hijacked agent, making stateful contextual policies a promising path toward safely deploying autonomous agents.

\section*{Acknowledgments}
We thank Marco Gruteser, Christoph Kern, Pierre Tholoniat, Julien Piet, Dan Ramage, Ananth Balashankar, Greg
Ganger, Hank Levy, and David Culler for their helpful feedback and ideas that
contributed to this work.

\bibliographystyle{ACM-Reference-Format}
\bibliography{paper}

\end{document}